\documentclass[letterpaper, 10 pt, journal, twoside]{IEEEtran} 

\IEEEoverridecommandlockouts                              

\usepackage{amsmath}
\usepackage{amssymb}
\usepackage{xcolor}
\usepackage{xspace}
\usepackage{float}
\usepackage{subfloat}
\usepackage{subfig}
\usepackage{bm}
\usepackage{enumitem}
\usepackage[font=small,labelfont=bf]{caption}
\usepackage[11pt]{moresize}
\usepackage{booktabs}
\usepackage{multirow}
\usepackage{multicol}
\usepackage{cite}
\usepackage{array,multirow,graphicx}
\usepackage{makecell}
\usepackage{authblk}
\usepackage{tikz}

\newcommand{\parlength}[1]{}
\newcommand{\myParagraph}[1]{{\bf #1.}}

\newcommand{\omitted}[1]{}

\title{
LAMP 2.0: A Robust Multi-Robot SLAM System for Operation in Challenging Large-Scale Underground Environments
}
\author{Yun Chang$^{1,*}$, Kamak Ebadi$^{2,*}$, Christopher E. Denniston$^{3}$, Muhammad Fadhil Ginting$^{2}$, Antoni Rosinol$^{1}$, Andrzej Reinke$^{4}$, Matteo Palieri$^{5}$, Jingnan Shi$^{1}$, Arghya Chatterjee$^{6}$, Benjamin Morrell$^{2}$, Ali-akbar Agha-mohammadi$^{2}$, Luca Carlone$^{1}$
\thanks{This work was supported by the Jet Propulsion Laboratory -- California Institute of Technology, under a contract with the National Aeronautics and Space Administration. This work was partially funded by the Defense Advanced Research Projects Agency (DARPA). \textcopyright  2022 All rights reserved. }
\thanks{$^*$Equal contribution. Corresponding Authors: yunchang@mit.edu, kamak.ebadi@jpl.nasa.gov} 
\thanks{$^{1}$Y.\,Chang, A.\,Rosinol, J.\,Shi, L.\,Carlone are with the Dept. of Aeronautics and Astronautics, Massachusetts Institute of Technology, United States.}
\thanks{$^{2}$K.\,Ebadi, M.\,Ginting, B.\,Morrell, A.\,Agha-mohammadi are with the NASA Jet Propulsion Laboratory -- California Institute of Technology, United States.} 
\thanks{$^{3}$C. E.\,Denniston is with Robotic Embedded Systems Laboratory, University of Southern California, United States. } 
\thanks{$^{4}$A.\,Reinke is with the University of Bonn, Germany.}
\thanks{$^{5}$M.\,Palieri is with the Dept. of Electrical And Information Engineering, Polytechnic University of Bari, Italy.}
\thanks{$^{6}$A.\,Chatterjee is with Dept. of Mechanical Engineering, Bangladesh University of Engineering and Technology, Bangladesh.}
\thanks{Digital Object Identifier (DOI): see top of this page.}
}
\usepackage{mathtools}
\usepackage{color}
\usepackage{graphicx}

\newcommand{\rev}[1]{{\color{black}#1}}
\newcommand{\revv}[1]{{\color{black}#1}}
\definecolor{green(ncs)}{rgb}{0.0, 0.62, 0.42}
\begin{document}
\maketitle

\begin{tikzpicture}[overlay, remember picture]
\path (current page.north east) ++(-6.0,-0.0) node[below left] {
This paper has been accepted for publication at RA-L. Please cite as:
};
\end{tikzpicture}
\begin{tikzpicture}[overlay, remember picture]
\path (current page.north east) ++(-2.8,-0.4) node[below left] {
Y. Chang, K. Ebadi, C. E. Denniston, M. F. Ginting, A. Rosinol, A. Reinke, M. Palieri, J. Shi, A. Chatterjee,
};
\end{tikzpicture}
\begin{tikzpicture}[overlay, remember picture]
\path (current page.north east) ++(-7.1,-0.8) node[below left] {
B. Morrell, A. Agha-mohammadi, and L. Carlone,
};
\end{tikzpicture}
\begin{tikzpicture}[overlay, remember picture]
\path (current page.north east) ++(-2.2,-1.2) node[below left] {
"LAMP 2.0: A Robust Multi-Robot SLAM System for Operation in Challenging Large-Scale Underground Environments,"
};
\end{tikzpicture}
\begin{tikzpicture}[overlay, remember picture]
\path (current page.north east) ++(-7.5,-1.6) node[below left] {
 in IEEE Robotics and Automation Letters, 2022.
};
\end{tikzpicture}

\begin{abstract}
Search and rescue with a team of heterogeneous mobile robots in unknown and large-scale underground environments requires high-precision localization and mapping. This crucial requirement is faced with many challenges in complex  and perceptually-degraded subterranean environments, as the onboard perception system is required to operate in off-nominal conditions (poor visibility due to darkness and dust, rugged and muddy terrain, and the presence of self-similar and ambiguous scenes). 
In a disaster response scenario and in the absence of prior information about the environment, robots must rely on noisy sensor data and perform Simultaneous Localization and Mapping (SLAM) to build a 3D map of the environment and localize themselves and potential survivors.
To that end, this paper reports on a multi-robot SLAM system developed by team CoSTAR in the context of the DARPA Subterranean Challenge. We extend our previous work, LAMP, by incorporating a single-robot front-end interface that is adaptable to different odometry sources and lidar configurations, 
a scalable multi-robot front-end to support inter- and intra-robot loop closure detection for large scale environments and multi-robot teams, 
and a robust back-end equipped with an outlier-resilient pose graph optimization based on Graduated Non-Convexity.
We provide a detailed ablation study on the multi-robot front-end and back-end, and assess the overall system performance in challenging real-world datasets collected across mines, power plants, and caves in the United States. 
We also release our multi-robot back-end datasets (and the corresponding ground truth), which can serve as challenging benchmarks for large-scale underground SLAM.
\end{abstract}
\section{Introduction\parlength{(1.25 pages)}}
\IEEEPARstart{S}{imultaneous} Localization And Mapping (SLAM)
 is a mature field of research, and there is a substantial body of literature dedicated to advancing SLAM algorithms and systems~\cite{cadena_tro_2016}. While there has been great progress in SLAM in urban and structured 
 environments, localization and mapping in extreme underground scenarios 
 has recently received increasing attention. In particular, the recent DARPA Subterranean Challenge~\cite{SubT}
 had the goal of
 developing 
 robotic systems capable of exploring and mapping complex underground environments, to support high-precision localization of elements of interest (e.g., survivors). This paper presents the design and implementation of a multi-robot lidar-centric SLAM system, as an enabling factor for autonomous exploration of challenging underground environments. 

\begin{figure}[t!]
\centering
	\includegraphics[width=0.98\columnwidth, trim= 0mm 120mm 0mm 0mm, clip]{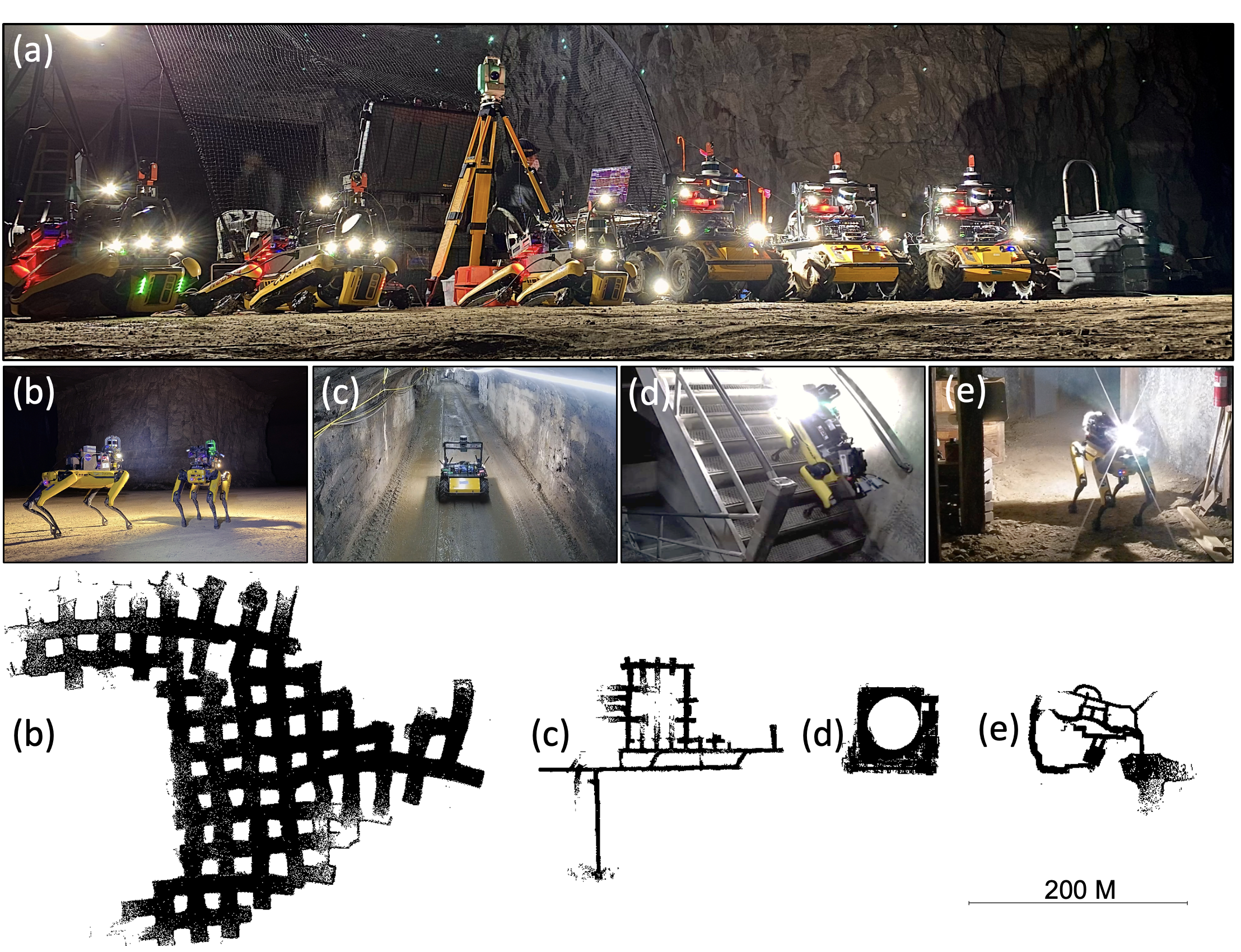} 
	\caption{\rev{(a) A heterogeneous team of robots undergoing initial calibration; 
	(b) Kentucky Underground dataset: 4 robots traversing over 6 km (combined) in the Kentucky Underground Storage, Wilmore, Kentucky;
	(c) Tunnel dataset: 2 robots covering 2.5 km in the NIOSH safety research mine, Pittsburgh, Pennsylvania;  
	(d) Urban dataset: 3 robots covering a multi-floor environment in the Satsop nuclear power plant, Elma, Washington;
	(e) Finals dataset: 4 robots mapping a heterogeneous course during the DARPA Subterranean Challenge finals in the Louisville Mega Cavern, Kentucky.
    \label{fig:cover} \vspace{-7mm}}}
\end{figure}

\myParagraph{Related work}\parlength{(0.75 page)}
We review related work on SLAM \emph{systems}, and particularly lidar-based SLAM systems for subterranean and perceptually degraded environments, and refer the reader to~\cite{cadena_tro_2016} for a broader survey on SLAM.
 Early efforts on SLAM in subterranean environments can be traced back to the works of Thrun et al.~\cite{thrun_icra_2003}
 and Nuchter et al.~\cite{nuchter_icra_2004}, 
which highlighted the importance of underground mapping and introduced early solutions involving a 
 cart pushed by a human operator, or teleoperated robots equipped with laser range finders to acquire volumetric maps of underground mines. Tardioli et al. \cite{tardioli2012underground, tardioli2018dumper} presented a SLAM system for exploration of underground tunnels using a team of robots. Zlot et al. \cite{zlot2012mine} presented a 3D SLAM system using a spinning 2D lidar and an industrial-grade MEMS IMU to map over 17 kilometers of an underground mine. 
The three-year-long DARPA Subterranean (SubT) challenge, which began in 2018, aimed at developing innovative approaches to rapidly map, navigate, and search underground environments with a team of autonomous robots. This worldwide competition has led to developing breakthrough technologies and capabilities for underground robotic operation~\cite{ebadi2020lamp,Palieri,DARE-SLAM,Roucek,Petrlik,Bouman_short,Williams,Kramer,Kratky,Miller,Ginting,Azpurua,Dang,Wisth,Khattak1,Queralta,Gross,Ohradzansky,Bayer,Tidd,Lindqvist,Rogers,Carter}.
Roucek et al.~\cite{Roucek} present a multi-robot heterogeneous system for robotic exploration of hazardous subterranean environments; \rev{the aerial platforms rely on Hector SLAM~\cite{Kohlbrecher}, and the ground robots use an ICP-based SLAM system~\cite{Pomerleau}.}
Similarly, Petrlik et al. \cite{Petrlik} present a system intended for aerial Search and Rescue (SAR) operations in underground settings. 
Kramer et al. \cite{Kramer} present a sparse pose graph-based visual-inertial SLAM system that relies on onboard lighting for exploration of the subterranean environments; 
the system uses multiple cameras, with significant overlap between their fields of view. The system’s front-end detects visual features and matches them to previously seen landmarks using BRISK features~\cite{BRISK}, and it uses IMU measurements to propagate the camera pose from the last optimized pose to the current frame time and predicts where matched landmarks should be observed in the new pose. The back-end uses batch optimization over a sliding window of recent camera frames and IMU measurements to estimate the system’s state. 

Azpurua et al. \cite{Azpurua} present a pipeline for semi-autonomous exploration of confined environments (e.g., pipes, underground tunnel systems) that includes both offline
photogrammetry for photorealistic map construction, and two parallel SLAM methods, a vision-based system based on RTAB-Map \cite{Labbe}, and a lidar-based system based on LeGO-LOAM \cite{Shan} for localization and mapping.
Dang et al. \cite{Dang,Khattak1} present a loosely-coupled multi-modal SLAM system based on fusion of LiDAR, visual, thermal, and inertial data to provide robust and resilient localization in underground
environments.
In a similar study, Wisth et al. \cite{Wisth} present a graph-based odometry system for mobile robotic platforms based on integration of lidar features with standard visual features and IMU data. The system is able to handle under-constrained geometry that affects lidar, or textureless areas that affects vision by using the best information available from each sensor modality without any hard switches.
Lajoie et al. present DOOR-SLAM \cite{DOOR-SLAM}, a fully distributed SLAM system which consists of two key modules, a pose graph optimizer combined with a distributed pairwise consistent measurement set maximization algorithm to reject spurious inter-robot loop closures, and a distributed SLAM front-end that detects inter-robot loop closures without exchanging raw sensor data.
Chang et al., present Kimera-Multi \cite{Kimera-Multi}, a fully distributed multi-robot system for dense metric-semantic SLAM. Each robot builds a local trajectory estimate and a local mesh. When two robots are within communication range, they initiate a distributed place recognition and robust pose graph optimization protocol. 

In our previous work \cite{ebadi2020lamp}, we presented \emph{Large-scale Autonomous  Mapping  and  Positioning} (LAMP), a pose-graph-based SLAM system developed in the context of the SubT Challenge. LAMP consisted of a front-end that performs lidar scan matching to obtain odometric estimates, and a back-end that performs pose graph optimization to obtain the best estimate of the robots' trajectories given odometry and loop closure measurements.
\rev{However, LAMP struggled with 
larger-scale long-duration operation in unknown subterranean environments
subject to perceptual aliasing.}

\myParagraph{Contribution}
This paper presents LAMP 2.0, a field-tested SLAM system for cooperative localization and mapping in unknown subterranean environments with a heterogeneous multi-robot team. 
The key contributions are:

\rev{\begin{enumerate}[leftmargin=*]
    \item A computationally efficient and outlier-resilient centralized multi-robot SLAM system that is adaptable to different input odometry sources, developed in the context of the DARPA Subterranean Challenge for operation in large-scale underground environments. The system includes the following improved modules:
    \begin{enumerate}
        \item A robust and scalable loop closure detection module that is able to handle and prioritize a rapidly growing number of loop closure candidates 
        and includes modern 3D registration techniques to improve the accuracy and robustness of the detected loop closures.
        \item An outlier-robust back-end based on 
        Graduated Non-Convexity~\cite{GNC} for pose graph optimization. 
    \end{enumerate}
    \item The open-source release of LAMP 2.0\footnote{https://github.com/NeBula-Autonomy/LAMP} and a multi-robot dataset\footnote{https://github.com/NeBula-Autonomy/nebula-multirobot-dataset} of subterranean environments, including the pose graph and point clouds of caves, mines, and abandoned power plants (Fig.~\ref{fig:cover}), 
    along with ground truth trajectories and maps based on 
    professionally surveyed data that can be used by the 
    SLAM community for evaluation of novel multi-robot localization and mapping solutions.
\end{enumerate}}


\section{LAMP 2.0\parlength{(2.25 pages)}} 
\begin{figure*}[t!]
    \centering
    \includegraphics[trim={5mm 5mm 5mm 5mm},clip, width=0.8\textwidth]{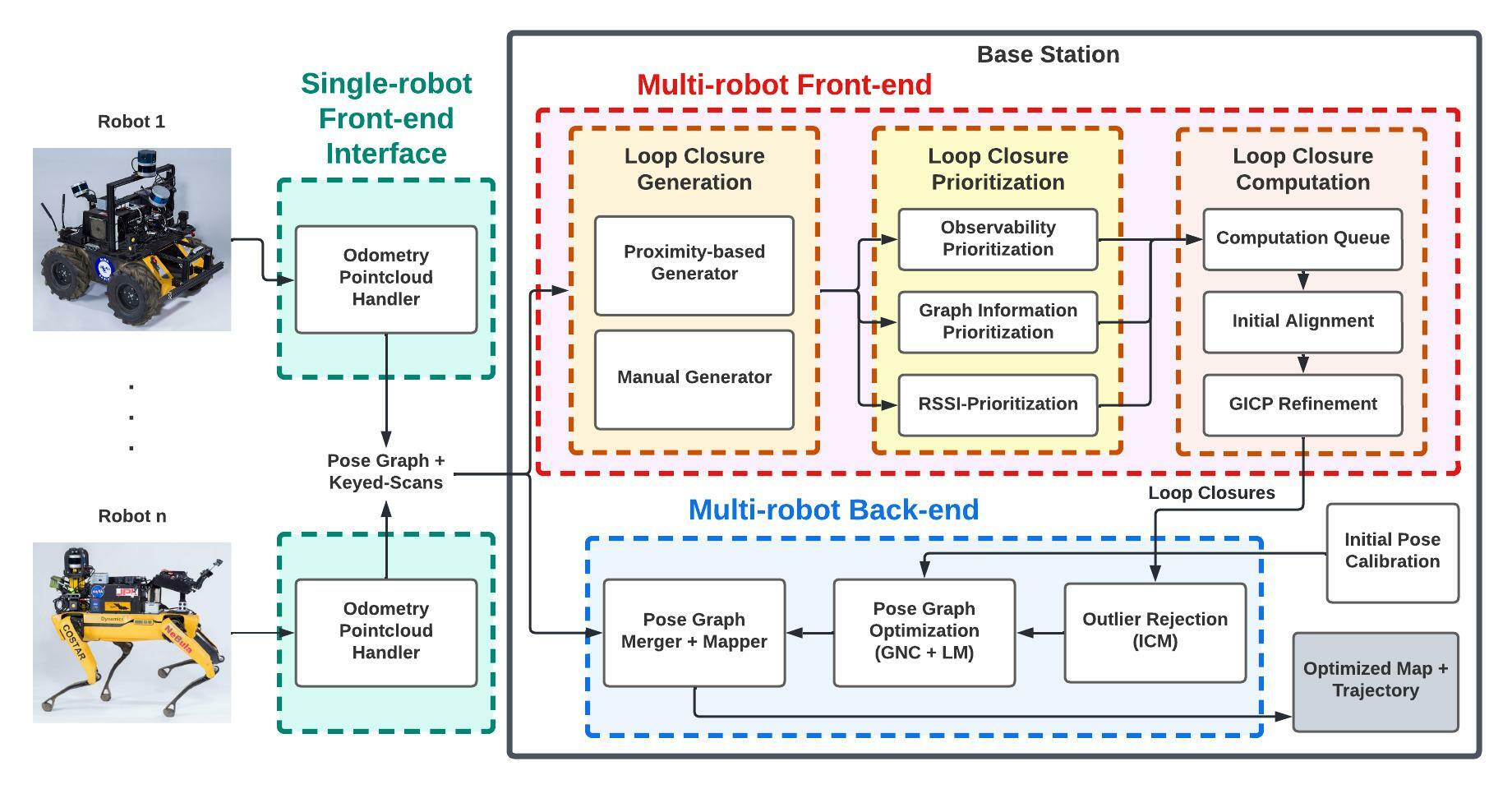}
    \caption{\revv{Architecture of LAMP 2.0, our multi-robot localization and mapping system, 
    which consists of a single-robot front-end interface to produce pose graph and keyed scans input (adaptable to different odometry sources and lidar configurations) on each robot, a multi-robot front-end for loop closure detection, and a multi-robot back-end for outlier-resilient pose graph optimization and map generation. }}\vspace{-5mm}
    \label{fig:architecture}
\end{figure*}
This section introduces LAMP 2.0, the SLAM system we developed in the context of the SubT Challenge for robust and scalable multi-robot SLAM in subterranean environments.
Fig.~\ref{fig:architecture} provides an overview of our system architecture.

The system consists of (i) a \emph{single-robot front-end interface} on each robot that takes in an odometry and point cloud stream and send to the centralized base station 
pose graph segments, which consists of odometry edges and nodes, and keyed scans, which are point clouds associated to odometry nodes, 
(ii) a \emph{multi-robot front-end}, running on a centralized base station, which receives the robots' local odometry and maps and performs multi-robot loop closure detection, and (iii) a \emph{multi-robot back-end}, that uses odometry (from all robots) and intra- and inter-robot loop closures from the front-end to perform a joint pose graph optimization; the multi-robot back-end is also executed at the central station and simultaneously optimizes all the robot trajectories. 
In the rest of this section, we highlight the core components of LAMP 2.0 that enable large-scale and robust, multi-robot localization and mapping in challenging underground environments.


\subsection{Initial Pose Calibration}
The sensors onboard each robot are assumed to be calibrated before operation (both intrinsic and extrinsic calibration).
Then, the initial pose calibration has the goal of establishing a common reference frame for all robots in the team. 
We achieve this by placing three reflective plates (fiducial markers) with known 3D coordinates on a gate at 
the entrance of the subterranean environment, and placing the robots in front of the gate.
The robots then begin scanning the environment using the onboard lidar scanners. The lidar points corresponding to the three reflective plates are then segmented into three clusters using Euclidean clustering. The lidar pose can then be computed by finding the centroids of the point clusters, and solving for the lidar pose that minimizes the distance between all triplets of cluster centroids, and the triplet of known fiducial markers' positions.

\rev{This gate calibration method is employed for compatibility with the SubT Challenge, but alternatives 
---such as the method in~\cite{tian2022kimera}---
can be also implemented to establish a common reference frame without initial pose calibration.}

\subsection{Adaptable Single-robot Front-end Interface}\label{sec:front-end-interface}
The single-robot front-end interface allows LAMP 2.0 to interface with single-robot front-ends with different odometry sources and different lidar configurations. 
As we will demonstrate in the experiments section, 
LAMP 2.0 is able to support in the same experiment, robots with different number and configuration of lidar sensors and possibly using different odometry front-ends, 
including LOCUS~\cite{LOCUS2} and Hovermap~\cite{Hovermap}.
In the following sections we detail how the interface processes the point-cloud and odometry inputs.

\myParagraph{Point cloud pre-processing} 
The front-end interface starts with a pre-processing step. 
First, any distortions in lidar scans caused by the robot motion are corrected using our Heterogeneous Robust Odometry (HeRO)~\cite{HeRO} system; 
HeRO computes a local state estimate, and the relative motion during the acquisition of a lidar scan is then corrected using the local state estimate. 
Next, given the extrinsic calibration between lidar scanners mounted on the robot, the undistorted point clouds are merged together into a single point cloud. 
To remove redundant and noisy points in the unified point cloud and to ensure that we retain a constant number of points 
regardless of environment geometry, point cloud density, and number of onboard lidar scanners,
we apply an adaptive voxelization filter~\cite{LOCUS2}.
The adaptive voxelization filter ensures that point clouds received from different lidar configurations are comparable in terms of size and density.
This improves the performance of the loop closure detection module
by reducing the computational load associated with loop closure transform estimation, 
and also reduces the computation and memory usage of the mapper on the centralized base station.

\myParagraph{Pose graph and keyed scans generation}
To represent the estimated robot trajectory, LAMP uses a pose graph formulation~\cite{cadena_tro_2016} 
where each node in the graph corresponds to an estimated pose (i.e., robot pose, or position of objects of interest in the scene), 
while each edge connecting two nodes in the graph encodes the relative motion or position measurement between the pair of nodes. 
It is crucial to prevent the pose graph from growing too large due to limited computation and memory available onboard each robot. 
Thus, we use a sparse pose graph with new nodes only added after the robot exceeds a motion threshold (e.g., 2 m in translation and 30 degrees of rotation). 
We refer to these robot pose nodes as the \emph{key nodes}. 
Each key node is associated with a \textit{keyed scan}, which is the pre-processed point cloud obtained at the corresponding time. 
The pose graph and the keyed scans are sent to the multi-robot front-end for loop closure detection, 
and to the back-end for pose graph optimization and 3D map generation.
\rev{When a robot moves out of communication range,
the pose graph and keyed scans are queued up and are then sent as a batch to the base-station 
when the communication is re-established.
We refer the reader to~\cite{nebula_short} for additional details on the networking setup.}
\subsection{Scalable Multi-robot Front-end}\label{sec:lidarLC}
In our SLAM architecture, the multi-robot front-end is in charge of intra- and inter-robot loop closure detection.
The ability to assert that a robot has returned to a previously visited location (i.e., a loop closure) is a crucial capability to reduce the accumulated error in the estimated robot trajectory. 
Moreover, loop closures are essential in detecting overlapping parts between local maps created by robots in the team, 
in order to fuse them into a geometrically consistent global map.
Our loop closure detection module consists of three steps: Loop closure generation, prioritization, and computation.
In the rest of this section, we describe the main components of LAMP 2.0's multi-robot front-end and highlight prioritization techniques that allow our system to scale to large-scale environments and large teams of robots.

\myParagraph{Loop closure generation}
In the generation step, loop closure candidates are generated
from nodes that lie within a certain Euclidean distance \rev{$d_{max}$} from the current node
in our pose-graph representation; \rev{$d_{max}$} is adaptive and is defined as 
\rev{$d_{max} = \alpha|n_{curr} - n_{cand}|$} based on the relative traversal between two nodes for the single-robot case, and 
\rev{$d_{max} = \alpha n_{curr}$} based on the absolute traversal for the multi-robot case, 
where $n_{curr}$ and $n_{cand}$ denote the \rev{sequentially assigned} index of the current node and the index of the candidate node respectively, and $\alpha$ is a fixed constant that decides the magnitude of \rev{$d_{max}$} with respect to the traversed distance (0.2 m in our experiments).
The design of the loop closure module is modular, 
such that other methods and environment representations (i.e.,  Bag-of-visual-words~\cite{DBOW2}, junctions extracted from 2D occupancy grid maps~\cite{DARE-SLAM}) can also be used for candidate generation.

\myParagraph{Loop closure prioritization}
The prioritization step inspects and adds the candidates to the queue for the computation step in an order
based on how likely they are to improve the trajectory estimate. 

Subterranean environments 
contain many self-similar and ambiguous scenes. While loop closure is crucial for map merging and drift reduction in the estimated robot trajectory, it is equally crucial to avoid closing loops in ambiguous areas with high degree of geometric degeneracy \cite{DARE-SLAM}, as it could lead to spurious loop closure detections, or inaccurate estimation of rigid-transformation between two corresponding nodes due to a lack of observability. 
Furthermore, loop closure detection in large-scale environments, and with multiple robots, 
becomes increasingly more challenging as the density of nodes in the pose graph, and subsequently the number of loop closure candidates, increases.
To overcome these challenges, our loop closure prioritization step includes three independent modules:

\rev{The \emph{Observability Prioritization} module prioritizes feature-rich areas, as they are more likely to produce more accurate loop closures.
To identify feature-rich scenes, similar to the works presented in~\cite{Tagliabue, Zhang, DARE-SLAM}, we rely on eigenvalue analysis of the information matrix of the pose estimate produced by point-to-plane ICP.

The \emph{Graph Information Prioritization} module prioritizes loop closures that can lead to a more substantial reduction of the trajectory error, given the structure of the graph.
We use a Graph Neural Network (GNN)~\cite{Zhou} to predict the outcome of a pose graph optimization triggered by a new loop closure.

The \emph{Receiver Signal Strength Indication (RSSI) Prioritization} module prioritizes loop closures 
based on known locations indicated by RSSI beacons. 
We refer the reader to~\cite{Frontend} for a more in-depth discussion and analysis of the prioritization module.}

\myParagraph{Loop closure computation}
The computation step calculates the relative transform for each candidate loop closure in the queue;
for each loop closure candidate, 
we implemented a two-stage pipeline to first find an initial alignment for the relative transform estimate 
using TEASER++~\cite{TEASER} or SAmple Consensus Initial Alignment (SAC-IA)~\cite{SAC-IA}),
then use the initial alignment to start the Generalized Iterative Closest Point (GICP) algorithm~\cite{GICP}
to obtain the refined relative transform and 
filter out matches that correspond to poor alignments (i.e., have a larger accumulated error than the threshold after the initial alignment or after GICP).
As presented in the experimental section, this two-stage pipeline significantly improves the accuracy and quality of loop closure detections, 
particularly because the GICP algorithm is prone to local minima.

\subsection{Robust Multi-robot Back-end}\label{sec:PGO}
In our SLAM architecture, the multi-robot back-end performs pose graph optimization to generate a globally consistent and drift-free 3D map of the environment.
We adopt a centralized multi-robot architecture, where a central base station receives the odometry measurements and keyed scans from each robot,
along with loop closures from the multi-robot front-end, 
and performs pose graph optimization to obtain the optimized trajectory. 
The optimized map is then generated by transforming the keyed scans to the global frame using the optimized trajectory.

\myParagraph{Outlier-robust pose graph optimization} 
The robust multi-robot back-end that receives odometry measurements from all robots in the form of odometry edges created by the single-robot front-end interface modules and inter- and intra-robot loop closures from the multi-robot front-end, and performs pose graph optimization to compute a globally consistent trajectory estimate for all robots in the team.

To safeguard against erroneous loop closures, 
our multi-robot back-end includes two outlier rejection options:
Incremental Consistency Maximization (ICM)~\cite{ebadi2020lamp},
which checks detected loop closures for consistency with each other and the odometry before they are added to the pose graph,
and Graduated Non-Convexity (GNC)~\cite{GNC}, 
which is used in conjunction with Levenberg-Marquardt 
to perform an outlier-robust pose graph optimization to obtain
both the trajectory estimates and inlier/outlier decisions on the loop closure not discarded by ICM.
Pose Graph Optimization and GNC are implemented using the Georgia Tech Smoothing and Mapping library (GTSAM)~\cite{GTSAM}.
Table~\ref{tab:lampvslamp2}, provides a summary of new capabilities and features implemented in LAMP 2.0, which extends our previous work~\cite{ebadi2020lamp}.

\begin{table}[t!]
\centering
\caption{LAMP vs. LAMP 2.0 capabilities by front-end (FE), back-end (BE) and overall system (Sys)}\label{tab:lampvslamp2}
\begin{tabular}{|c|c|c|c|}
\hline
\parbox[t]{2mm}{\multirow{1}{*}{\rotatebox[origin=c]{90}{}}}   & Capability & LAMP & LAMP 2.0 \\ \Xhline{2\arrayrulewidth}
\parbox[t]{2mm}{\multirow{5}{*}{\rotatebox[origin=c]{90}{FE}}}
         & Proximity-based loop closure generation       & X & X \\ \cline{2-4}
         & Adaptive Proximity-based loop gen             &   & X \\ \cline{2-4}            
         & Loop closure prioritization                   &   & X \\ \cline{2-4}
         & GICP loop closure pose estimation             & X & X \\ \cline{2-4}
         & Two stage loop closure pose estimation        &   & X \\ \cline{2-4}
\Xhline{2\arrayrulewidth}
\parbox[t]{2mm}{\multirow{3}{*}{\rotatebox[origin=c]{90}{BE}}}
         & LM / GN pose graph optimization               & X & X \\ \cline{2-4}
         & ICM outlier rejection                         & X & X \\ \cline{2-4}
         & GNC in conjunction with LM / GN               &   & X \\ \cline{2-4}
\Xhline{2\arrayrulewidth}
\parbox[t]{2mm}{\multirow{2}{*}{\rotatebox[origin=c]{90}{Sys}}}
         & Adaptable to different lidar configs          &   & X \\ \cline{2-4}
         & Adaptable to different odometry input         &   & X \\ \cline{2-4}
\Xhline{2\arrayrulewidth}
\end{tabular}
\vspace{-6mm}
\end{table}

\section{Experiments}
This section showcases the performance of LAMP 2.0 and provides experimental results to highlight the new features and improvements from the previous version~\cite{ebadi2020lamp}.
Section~\ref{sec:exp_hw_setup} describes our hardware setup.
Section~\ref{sec:exp_data} describes the dataset we collected, open-sourced, and used for evaluation. 
Section \ref{sec:exp_components} evaluates and discusses the performances of selected components in the LAMP 2.0 multi-robot front-end and back-end.
Section \ref{sec:exp_full} reports on the performance of the overall LAMP 2.0 system, including its performance in the three events of the SubT challenge. 

\subsection{Hardware Setup} \label{sec:exp_hw_setup}
\myParagraph{Robots} 
In the experiments we present in this paper, the datasets are collected with two types of 
robots: Husky and Spot robots.
The Husky is a wheeled platform and is equipped with three Velodyne lidars and a Hovermap~\cite{Hovermap}. 
The odometry input to LAMP 2.0 is either provided by LOCUS~\cite{LOCUS2} or the Hovermap~\cite{Hovermap}.
The Spot robot is a quadruped platform equipped with either a single lidar or a Hovermap. 
Similarly to the Husky, the odometry input to LAMP is either provided by the Hovermap or LOCUS~\cite{LOCUS2}.
Note that not only do the odometry inputs to the single-robot front-end interface vary (between Hovermap and LOCUS), 
but also the point cloud inputs vary in size and density 
between the Spot robots with a single lidar, Spot robots with a Hovermap, 
Husky robots with three lidars, and Husky robots with a Hovermap.

\myParagraph{Base station}
During the SubT Challenge, the base was a portable workstation 
with an AMD Ryzen Threadripper 3990x processor with 64 cores.  
The experiments in this paper ran on a laptop with  Intel i7-8750H processor with 12 cores.

\subsection{Datasets} \label{sec:exp_data}
We evaluate LAMP 2.0 on four datasets collected by Team CoSTAR.
The first dataset is the {\bf Tunnel Dataset}, which was collected in the NIOSH Safety Research Coal Mine in Pittsburgh, PA,
and includes two Huskies traversing up to 2.5 km combined, in a coal mine 
that consists of mostly featureless narrow tunnels.
The second dataset is the {\bf Urban Dataset}, which was collected in the  Satsop abandoned nuclear power plant in Elma, WA, 
and includes two Huskies and a Spot traversing up to 1.5 km combined in an abandoned nuclear power plant 
consisting of  a two-floor environment with open areas, small rooms, narrow passageways, and stairs.
The third dataset is the {\bf Finals Dataset}, which was collected during the second preliminary run of the SubT finals and includes three Spots
and a Husky traversing up to 1.2 km combined in the DARPA-built course including tunnel, cave, and urban-like environments.
The last dataset is the {\bf Kentucky Underground Dataset (KU)}, which was collected by Team CoSTAR in the
Kentucky Underground Storage in Wilmore, KY, and includes four Huskies traversing up to 6 km combined
in a limestone mine, which consists of large 10-20 m wide tunnels.
The maps and sample scenes from the four datasets are shown in Fig.~\ref{fig:cover}.

For each dataset, we used a surveyed global map to generate an odometry ground-truth through scan-to-map localization~\cite{LOCUS2}. 
To test the performance of multi-robot loop closure detection, we generate sets of ground-truth 
loop closure pairs from the ground-truth trajectory, including the correct transform,
and also a set of false loop closures to stress the robustness of the system.

We have prepared these datasets, in the format of pose graphs and keyed scans,
and released them open-source to promote further research on multi-robot loop closure detection and robust pose graph optimization.

\subsection{Component Evaluation} \label{sec:exp_components}
This section focuses on the evaluation of the multi-robot front-end and the multi-robot back-end,
and highlights the new features in these two modules 
compared to our previous system in~\cite{ebadi2020lamp},
in order to assess their impact on our new LAMP 2.0 system.

We will first demonstrate the improvements to loop closure detection accuracy 
with the two-stage loop closure computation module. 
Then, we will present an ablation study with different outlier rejection configurations
and highlight the performance of GNC.
Lastly, we will evaluate the multi-robot front-end and back-end in conjunction, 
and show that with the new features in the front-end and the back-end, 
we are able to include more accurate loop closures in our pose graph 
and obtain a better trajectory and map estimate.
We refer the reader to~\cite{Frontend} for detailed experiments and analysis of the loop closure prioritization module.

\myParagraph{Two-stage loop closure computation}
We evaluate the results from the loop closure computation step for different types of initializations for ICP.
Previously, LAMP used an odometric initialization approach~\cite{ebadi2020lamp};
on the other hand, LAMP 2.0 uses a two-stage approach and uses TEASER++~\cite{TEASER} or SAC-IA~\cite{SAC-IA}
for initial alignment (stage 1) and ICP for final refinement (stage 2).
Table~\ref{tab:lc-ablation} shows the recall and false-positive rate, 
along with the mean translation and rotation error of the different initialization 
methods for ICP.
In our experiments, we set the SAC cumulative error threshold to 32 m, 
the max number of iterations of SAC to 500, the ICP cumulative error threshold to 0.9 m,
and the max number of ICP iterations to 200.
The recall is the percentage of correct loop closures that passed SAC and ICP 
(i.e. had a lower SAC error than the threshold after initial alignment 
and had a lower ICP error than the threshold after ICP registration)
and the false positive rate is the percentage of false loop closures that passed SAC and ICP.
The mean and translation errors are computed using the correct loop closures that passed SAC and ICP
against the ground truth trajectory.
The odometry initialization typically has a better recall (i.e., more loop closures are found) in narrower environments, while  TEASER++ and SAC-IA have better recall in wider environments (e.g., Urban and KU).
More importantly, the newly available TEASER++ and SAC-IA initializations 
lead to  a significant decrease in false-positive rate and mean translation and rotation error. 
\rev{The lower false-positive rate when using TEASER++ or SAC-IA is largely due to 
the additional filtering of poor alignments
that do not have a sufficient number of inliers.
LAMP 2.0 shows a decrease in the number of incorrect loop closures and increased pose accuracy for the correct loop closures.} 

\setlength{\tabcolsep}{2pt}
\begin{table}
\centering
\caption{Comparison of different initialization methods for ICP-based loop closure relative pose estimation }\label{tab:lc-ablation}
\begin{tabular}{cl cccc}
\toprule
& & \multicolumn{4}{c}{Initialization Methods} \\
\midrule
& & GT & OdomRot~\cite{ebadi2020lamp} & TEASER++ & SAC-IA \\
\midrule
\multirow{4}{*}{$\uparrow$ Recall (\%)}
& Tunnel & 90.8 & \textbf{93.9} & 76.6 & 81.9 \\
& Urban & 90.5 & 78.2 & 78.2 & \textbf{79.6} \\
& Final & 89.2 & \textbf{83.4} & 68.4 & 57.0 \\
& KU & 29.0 & 11.3 & \textbf{18.5} & 17.7 \\
\midrule
\multirow{4}{*}{$\downarrow$ F/P (\%)}
& Tunnel & 2.0 & 2.4 & \textbf{1.2} & 1.4 \\
& Urban & 0.8 & 1.6 & \textbf{0.6} & 1.0 \\
& Final & 1.2 & 7.4 & \textbf{0.6} & 1.0 \\
& KU & 0.4 & \textbf{0.0} & 0.2 & 0.0 \\
\midrule
\multirow{4}{*}{$\downarrow$ Mn Err. (m/deg)}
& Tunnel & 0.09/0.86 & 0.86/8.02 & 0.71/10.82 & \textbf{0.67/9.63} \\
& Urban & 0.44/1.47 & 1.89/1.98 & \textbf{0.38/1.36} & 0.52/1.6 \\
& Final & 0.06/1.11 & 1.65/6.43 & \textbf{0.32/2.39} & 0.6/2.99 \\
& KU & 0.26/0.96 & 0.82/1.34 & \textbf{0.29/1.45} & 0.29/1.48 \\
\end{tabular}
\vspace{-5mm}
\end{table}

\begin{figure}[t!]
\centering
\subfloat[Inliers translation error] {\includegraphics[trim={20, 5, 45, 40}, clip, width=0.46\columnwidth]{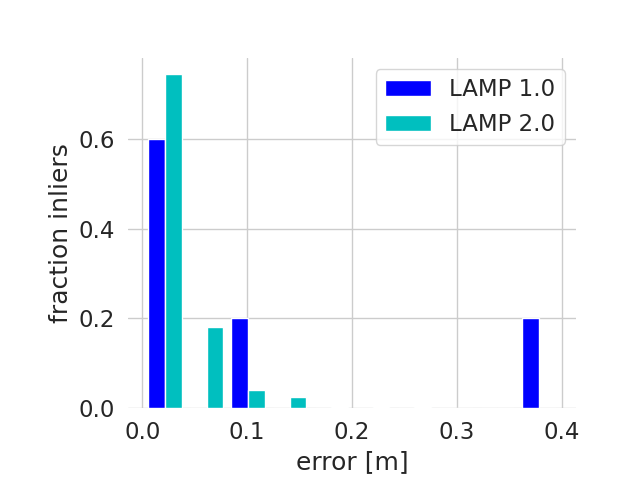}}
\hfill
\subfloat[Inliers rotation error] {\includegraphics[trim={20, 5, 45, 40}, clip, width=0.46\columnwidth]{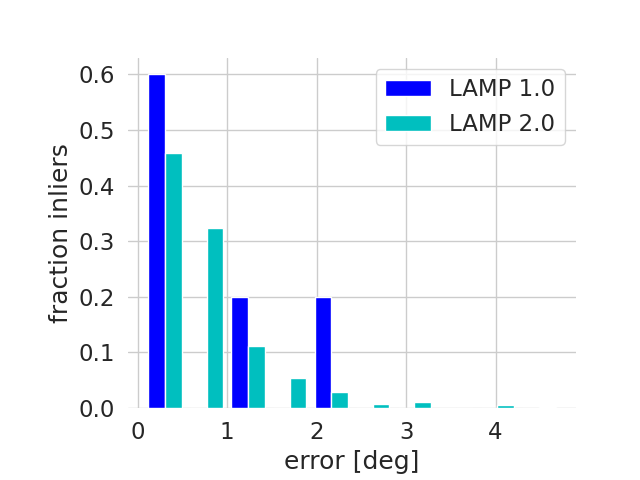}}
\caption{\rev{Pose error of the loop closure inliers 
from LAMP 1.0 (Old Loop Closure module + ICM) compared to LAMP 2.0 (New Loop Closure module + GNC) for the Finals dataset.} \vspace{-5mm}}
\label{fig:lc-comparison}
\end{figure}

\begin{figure*}[t!]
\centering
\includegraphics[trim=40 20 20 20, clip, width=0.95\textwidth]{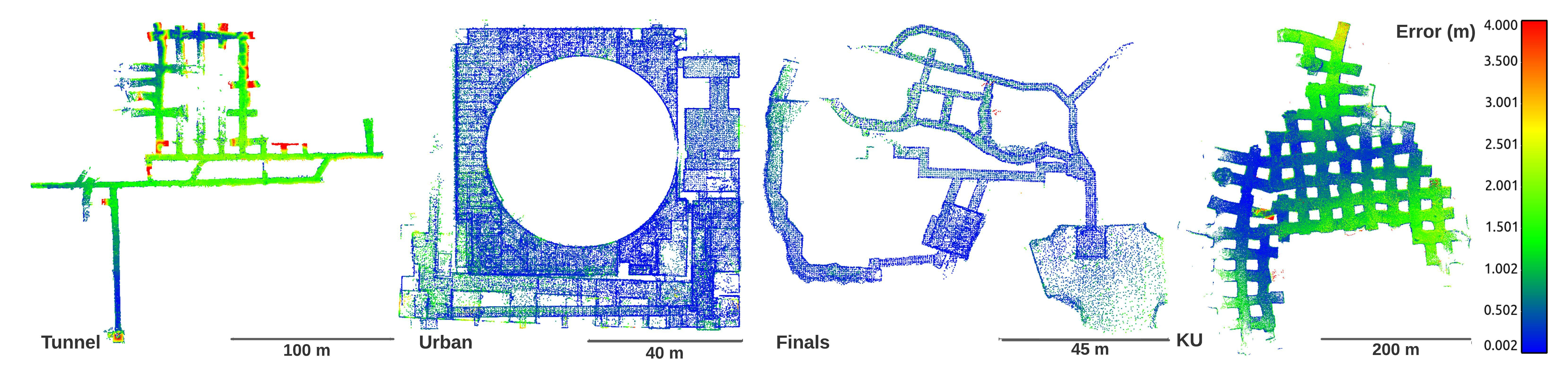}
\caption{Final map error in meters of the DARPA Subterranean Challenge finals course for the Tunnel, Urban, Finals, and Kentucky Underground (KU) datasets. \vspace{-7mm}} 
\label{fig:mapping-results}
\end{figure*} 

\begin{figure}
\centering
\subfloat[Tunnel ATE]{\includegraphics[trim={25 5 35 40}, clip, width=0.48\columnwidth]{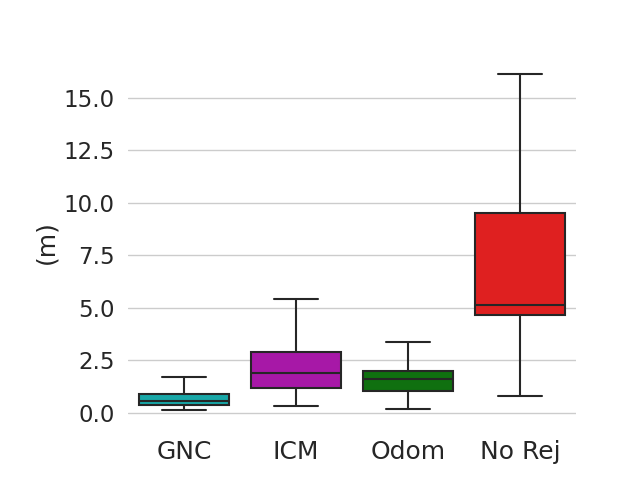}}
\hfill
\subfloat[Urban ATE]{\includegraphics[trim={25 5 35 40}, clip, width=0.48\columnwidth]{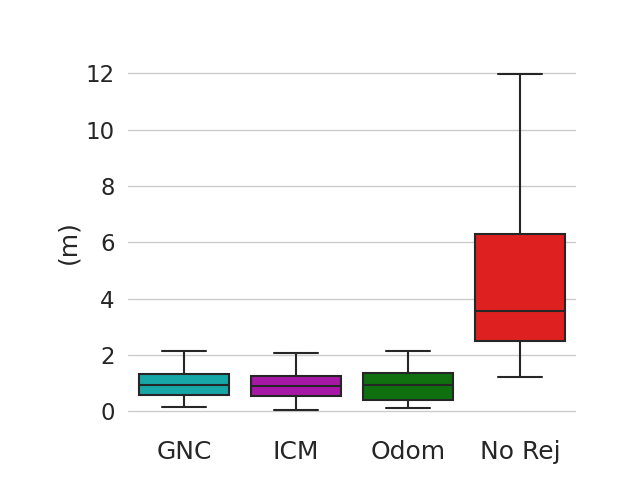}}
\hfill
\subfloat[Finals ATE]{\includegraphics[trim={25 5 35 40}, clip, width=0.48\columnwidth]{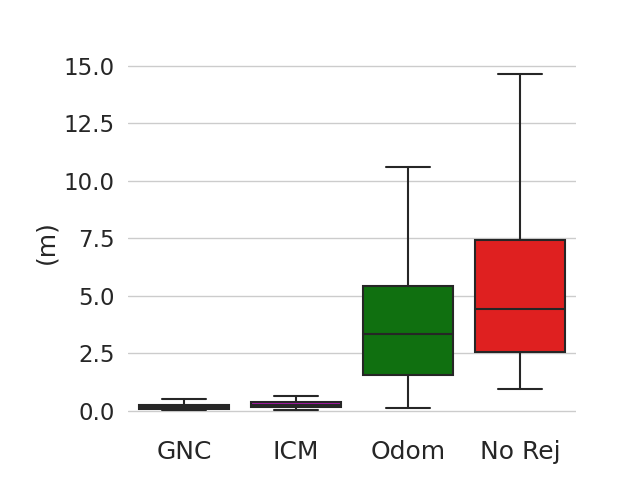}}
\hfill
\subfloat[KU ATE]{\includegraphics[trim={25 5 35 40}, clip, width=0.48\columnwidth]{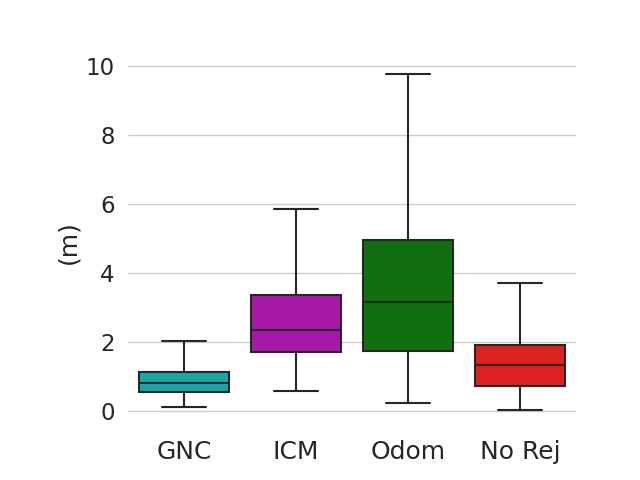}}
\caption{Comparison of trajectory ATE (across the multi-robot trajectory) for ICM and GNC compared to the cases with no loop closure or no outlier rejection. \vspace{-5mm}}
\label{fig:or-ablation-blxplt}
\end{figure}

\myParagraph{Outlier-robust pose graph optimization}
We demonstrate the capabilities of our outlier-robust pose graph optimization module with an ablation study,
especially focusing on the performance of the newly added GNC option 
as compared to ICM in~\cite{ebadi2020lamp}.
In Fig. \ref{fig:or-ablation-blxplt} we compare the average final trajectory error
with GNC or ICM as an outlier rejection method; we also report trajectory errors for the case when no outlier rejection is performed (``No Rej'') and when no loop closure is detected (``Odom''). While both GNC and ICM largely improve the trajectory estimate 
relative to the no loop closure or no outlier rejection case,  GNC in general is more robust to outliers and gives better trajectory estimates, 
as especially evident in the longer Tunnel and Kentucky Underground datasets \rev{where ICM failed to reject some of the outliers in the first and was too conservative in the other.} 

Table~\ref{tab:lc-numbers} shows the number of loop closure candidates generated ("Generated"), 
the number of loop closures that passed ICP ("Verified"), and the number of ICM or GNC inliers ("Inliers").
Recall that LAMP 1.0~\cite{ebadi2020lamp} has only a fixed radius for candidate generation, 
no prioritization step, and uses an odometric ICP initialization for loop closure pose estimation and  ICM as the outlier rejection method; 
on the other hand, the proposed  LAMP 2.0 system 
includes candidate generated with an adaptive radius, the prioritization step, and uses SAC-IA/TEASER++ for ICP initialization and GNC as an outlier rejection method.
With the new loop closure module and prioritization, 
even though we generate fewer candidates,
we are able to find a larger number of verified and inlier loop closures, 
showing the validity of the adaptive radius, 
and that prioritization does prioritize loop closures of higher ``quality''.
Notice also that GNC is able to perform well with more than 80\% outlier loop closures detected.
Fig.~\ref{fig:lc-comparison} compares the translation and rotation errors (with respect to ground truth) of loop closures considered inliers by LAMP 1.0~\cite{ebadi2020lamp} and LAMP 2.0. The figure shows how LAMP 2.0 \rev{not only produces a significantly larger number of inlier loop closures but also that the detected inliers are better biased towards lower errors compared to LAMP 1.0}, confirming that the prioritization and initial alignment (Table~\ref{tab:lc-ablation}), together with the outlier rejection (Fig.~\ref{fig:or-ablation-blxplt}) lead to an increased number of more accurate loop closures.

\setlength{\tabcolsep}{5pt}
\begin{table}
\centering
\caption{Amount of loop closures at the different stages in the multi-robot frontend and backend.}\label{tab:lc-numbers}
\begin{tabular}{cl ccc}
\toprule
& & \# Generated & \# Verified & \# Inliers \\
\midrule
\multirow{2}{*}{Tunnel}
& LAMP 1.0 & 22206 & 3032 & 52 \\
& \textbf{LAMP 2.0} & \textbf{9755} & \textbf{5656} & \textbf{1645} \\
\midrule
\multirow{2}{*}{Urban}
& LAMP 1.0 & 17742 & 497 & 237 \\
& \textbf{LAMP 2.0} & \textbf{5356} & \textbf{1505} & \textbf{284} \\
\midrule
\multirow{2}{*}{Final}
& LAMP 1.0 & 12559 & 681 & 6 \\
& \textbf{LAMP 2.0} & \textbf{3856} & \textbf{1484} & \textbf{393} \\
\midrule
\multirow{2}{*}{KU}
& LAMP 1.0 & 22616 & 14 & 11 \\
& \textbf{LAMP 2.0} & \textbf{22344} & \textbf{885} & \textbf{197} \\
\end{tabular}
\vspace{-7mm}
\end{table}

\begin{figure}[t]
\centering
\subfloat[KU single robot input]{\includegraphics[trim={0 50 0 60}, clip, width=0.45\columnwidth]{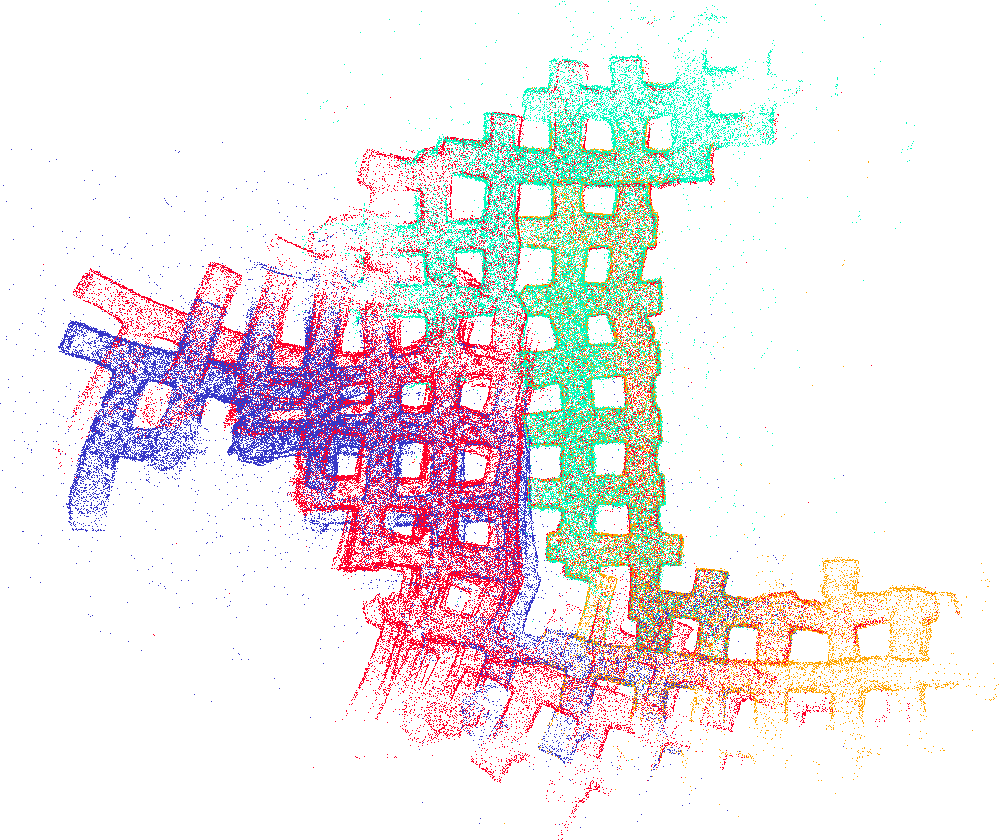}}
\hfill
\subfloat[KU LAMP 2.0 output]{\includegraphics[trim={0 50 0 60}, clip, width=0.45\columnwidth]{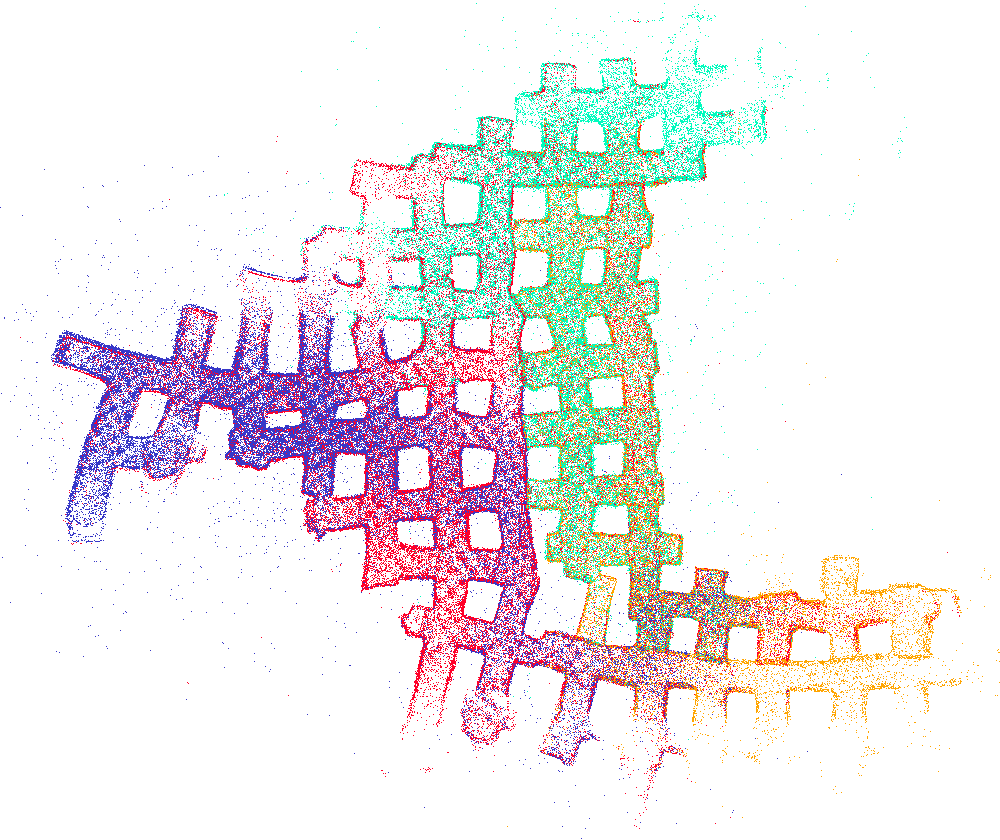}}
\caption{\rev{Single-robot map (no loop closures) compared with LAMP 2.0 map (with inter- and intra- robot loop closures).} \vspace{-6mm}}
\label{fig:ku_comparison}
\end{figure}

\subsection{System Evaluation}\label{sec:exp_full}
In this section, we provide end-to-end results of the full LAMP 2.0 system.
We play back the data in real-time to the base-station; the data is processed in the same order and rate as it was generated in the field \revv{and the results are collected at the end of data-stream ($\sim 1$ hr).} 
\revv{The odometry input is also consistent across the tests for fair comparison.}
In Table~\ref{tab:full-comparison}, we show the improvement in average trajectory error 
of LAMP 2.0 compared to LAMP 1.0 \revv{and LAMP 2.0 without inter-robot loop closures} for each robot on the four datasets.
LAMP 2.0 achieves minimal errors (below 2 m) 
with trajectory lengths of up to 2.2 km.
\rev{Fig.~\ref{fig:ku_comparison} illustrates the impact of inter- and intra- robot loop closures.}
Finally, we present the full multi-robot mapping results
by showing the LAMP 2.0 point cloud map colored by the cloud-to-cloud error
against the ground truth map for the four datasets in Fig.~\ref{fig:mapping-results}.
We are able to achieve map errors below 4 m
in these different large-scale and challenging environments.

\setlength{\tabcolsep}{2pt}
\begin{table}[t!]
\centering
\caption{Comparison of LAMP 2.0 against LAMP 1.0 }\label{tab:full-comparison}
\begin{tabular}{cl cccc}
\toprule
& \multirow{ 2}{*}{Robot} & Traversed  & LAMP 2.0 & \rev{Single Robot} & LAMP 1.0  \\
&  &  [m] & ATE [m] & ATE [m] & ATE[m] \\
\midrule
\multirow{2}{*}{Tunnel}
& husky3 & 1194 & \textbf{0.65} & 0.83 & 1.07 \\
& husky4 & 1362 & 0.72 & \textbf{0.63} & 1.44 \\
\midrule
\multirow{3}{*}{Urban}
& husky1 & 612 & \textbf{0.79} & 0.87 & 0.95 \\
& husky4 & 416 & \textbf{0.76} & 0.79 & 0.78 \\
& spot1 & 502 & 1.31 & 1.46 & \textbf{0.99} \\
\midrule
\multirow{4}{*}{KU}
& husky1 & 2204 & 1.01 & \textbf{0.9} & 5.34 \\
& husky2 & 1526 & \textbf{0.71} & 0.75 & 2.85 \\
& husky3 & 1678 & \textbf{1.31} & 1.33 & 2.11 \\
& husky4 & 896 & \textbf{0.69} & 0.86 & 5.49 \\
\midrule
\multirow{4}{*}{Final}
& husky3 & 72 & \textbf{0.16} & 0.16 & 0.56 \\
& spot1 & 430 & \textbf{0.2} & 0.37 & 0.37 \\
& spot3 & 484 & \textbf{0.21} & 0.38 & 0.55 \\
& spot4 & 238 & \textbf{0.15} & 0.23 & 0.62 \\
\end{tabular}
\vspace{-6mm}
\end{table}




\section{Conclusion}
In this paper we presented LAMP 2.0, a centralized multi-robot SLAM system, developed in the context of the DARPA Subterranean Challenge, which provides a robust estimate of the trajectories of multiple robots and constructs a point cloud map using 3D lidar data.
LAMP 2.0 includes a single-robot front-end interface that is adaptable to different odometry inputs and robots with different lidar configurations.
The multi-robot back-end is based on a modular ICP initialization framework to improve the convergence of the algorithm in ambiguous settings, and uses a loop closure prioritization module to deal with the growing number of loop closure candidates in large-scale environments. Then the multi-robot back-end runs an outlier-robust pose graph optimization to estimate the trajectories of all robots in the team. While \rev{LAMP 2.0} performed well in our experiments with up to four robots, its centralized architecture may not scale to large robot teams. Extending LAMP to a fully distributed system in underground domains with intermittent inter-robot communications is an interesting direction for future work.


\bibliographystyle{IEEEtran}
\bibliography{references}

\end{document}